\pdfoutput=1
\documentclass[11pt]{article}

\usepackage[]{acl}

\usepackage{times}
\usepackage{latexsym}
\usepackage{amsmath}
\usepackage{acronym}
\usepackage{cprotect}
\usepackage{lipsum}% http://ctan.org/pkg/lipsum
\usepackage{graphicx}% http://ctan.org/pkg/graphicx

\usepackage[norelsize, linesnumbered, ruled, lined, boxed, commentsnumbered]{algorithm2e}

\usepackage[T1]{fontenc}
\usepackage[utf8]{inputenc}

\usepackage{microtype}
\usepackage{enumitem}
\setlist{noitemsep,topsep=0pt,leftmargin=10pt}

\SetKwComment{Comment}{/* }{ */}
\title{Language Agnostic Code-Mixing Data Augmentation by Predicting Linguistic Patterns}

\author{Shuyue Stella Li \and Kenton Murray \\
         Center for Language and Speech Processing, Johns Hopkins University \\ 
         \texttt{sli136,kenton@jhu.edu}}

\begin{document}
\maketitle
\newacro{cmi}[CMI]{Code-Mixing Index}
\newacro{scm}[SCM]{Synthetic Code-Mixing}
\newacro{ncm}[NCM]{Natural Code-Mixing}
\newacro{pos}[POS]{Part-of-speech}
\newacro{nlp}[NLP]{natural language processing}
\newacro{sa}[SA]{Sentiment analysis}
\newacro{plm}[PLM]{Pre-trained Language Models}
\begin{abstract}
%Code-mixing, also called code-switching, is the linguistics phenomenon where in casual settings, multilingual speakers mix words from different languages in one sentence. Due to its spontaneous nature, code-mixing is extremely low-resource, especially in text data as opposed to its more common occurring speech settings. In many low-resource NLP settings, data augmentation has been shown to improve downstream tasks.
In this work, we focus on intrasentential code-mixing and propose several different Synthetic Code-Mixing (SCM) data augmentation methods that outperform the baseline on downstream sentiment analysis tasks across various amounts of labeled gold data. 
Most importantly, our proposed methods demonstrate that strategically replacing parts of sentences in the matrix language with a constant mask significantly improves classification accuracy, motivating further linguistic insights into the phenomenon of code-mixing. We test our data augmentation method in a variety of low-resource and cross-lingual settings, reaching up to a relative improvement of 7.73\% on the extremely scarce English-Malayalam dataset.
We conclude that the code-switch pattern in code-mixing sentences is also important for the model to learn. Finally, we propose a language-agnostic SCM algorithm that is cheap yet extremely helpful for low-resource languages.
\end{abstract}

\section{Introduction}
\label{sec:introduction}

\begin{figure}
\centering
\includegraphics[width=0.47\textwidth]{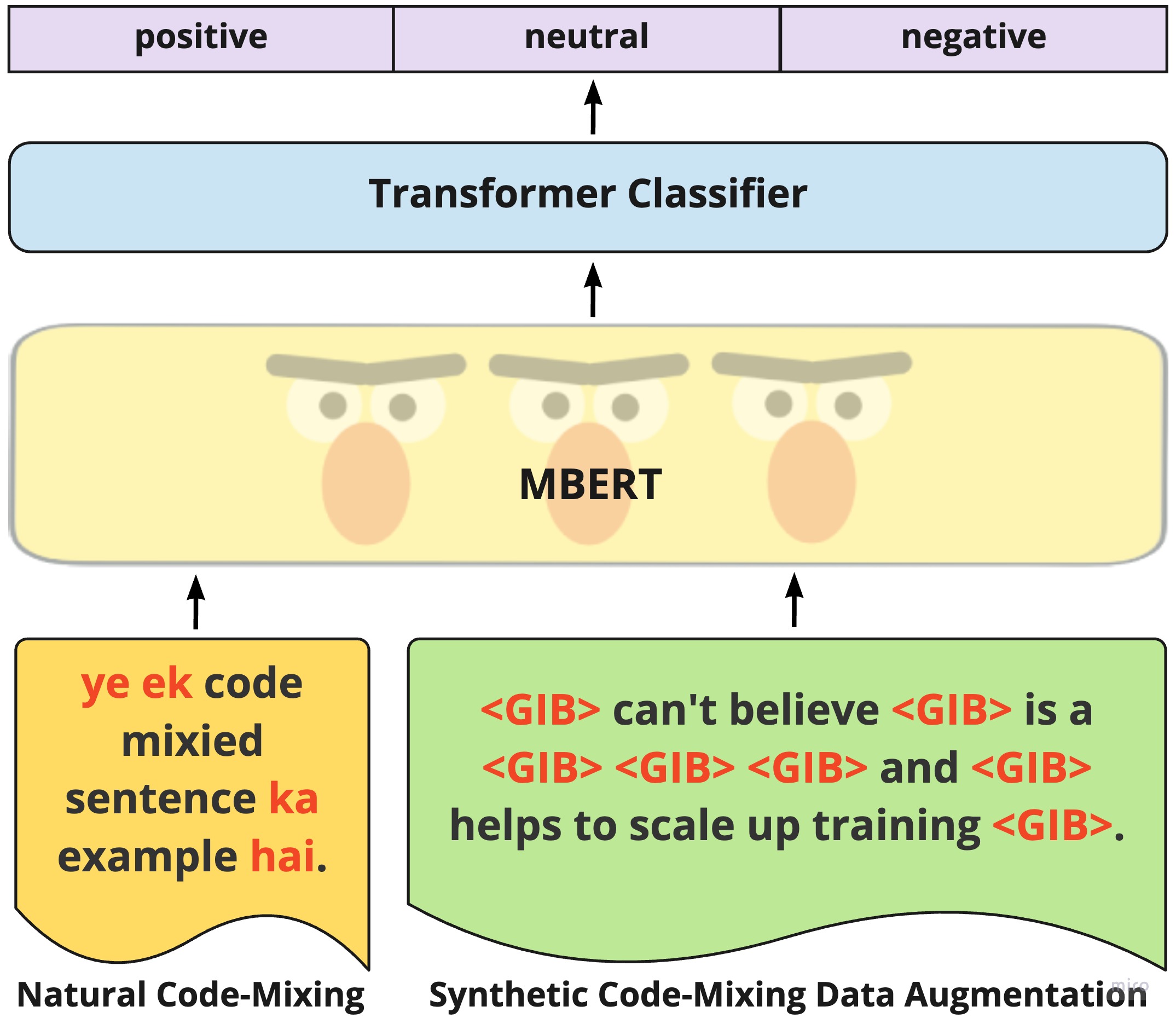}
\caption{SCM Data Generation Model Pipeline.\\ Large amounts of synthetic data are combined with limited natural data to finetune a multilingual PLM \cite{devlin2018mbert} with a transformer classifier layer for sentiment analysis.}\label{fig:modelpipeline}\vspace{-5mm}
\end{figure}

Code-mixing sentences have complex syntactic structure and a large vocabulary across languages. It is difficult for someone not fluent in both languages to understand a code-mixed conversion. Due to its spontaneous nature, code-mixing text data is hard to collect and therefore extremely low-resource. Most text-based code-mixing happens in casual settings such as blogs, chats, product ratings and comments, and most predominantly on social media. A model for \ac{nlp} tasks for code-mixed languages is necessary, but current \ac{nlp} studies on social media texts focus mostly on English \cite{farzindar2015natural, coppersmith2018natural, hodorog2022machine, oyebode2022covid}. However, the fact that English has become the \textit{lingua franca} in most social media apps can be helpful in cross-lingual generalization of code-mixing languages with one of the language being English \cite{choudhury_2018}. A recent survey on code-mixing datasets for any downstream tasks, found that 84\% of the code-mixing pairs contain English \cite{jose2020survey}. Therefore, having a model for English-\textbf{\textit{X}} (any other language) code-mixing languages is crucial for future \ac{nlp} research on code-mixing.

\ac{sa} is an important research area in social media \ac{nlp}, which learns to predict the sentiment of a piece of informal text, often in the form of a few sentences or a paragraph. \ac{sa} models are widely used in social media for social media monitoring \cite{ortigosa2014sentiment}, video/post suggestion, and commercially for analyzing customer feedback, business trend, etc \cite{drus2019sentiment}. In the code-mixing setting, there is also a need for \ac{sa} tools as the switching and choice of language for a multilingual speaker contains higher level implications that are valuable for downstream needs \cite{kim2014sociolinguistic}. Code-mixing sentences are composed of multiple languages, making multilingual \ac{plm} \cite{devlin2018mbert} the natural choice for the starting point to explore NLP tasks in this new domain. However, as we will discuss in Section \ref{sec:plm}, there are limitations on the ability of multilingual \ac{plm}s to be directly adopted for code-mixing languages.

As shown in Figure \ref{fig:modelpipeline}, we attempt to solve the domain mismatch between multilingual \ac{plm} and extremely low-resource, unseen code-mixing data by introducing a \ac{scm} data augmentation pipeline for code-mixing social media sentiment analysis. In our proposed data augmentation pipeline, large \ac{scm} data is combined with limited \ac{ncm} data to fine-tune a pre-trained language mode with a classification layer.

We introduce a low-cost, language-agnostic, and label-preserving algorithm to mass produce synthetic code-mixing sentences. 
In this work, our contributions include:
\begin{itemize}
    \item Our synthetic English-Hindi code-mixing data augmentation technique is shown to be extremely helpful for \ac{sa} across a variety of model complexity and low-resource levels.
    \item We investigate the linguistic nature of code-mixing from a computational perspective, concluding that the models learn from the form of code-mixing more so than from the semantics of individual constituents.
    \item We introduce a low-cost, \emph{language-agnostic}, and label-preserving algorithm that allows for the rapid production of a universal corpus for code-mixing sentences. We demonstrated improvement over all the language pairs and a 7.73 percent improvement in the weighted f1 score on extremely low-resource languages.
\end{itemize}

\section{Related Work}
\label{sec:relatedwork}
\subsection{Code-Mixing}
The earliest studies on code-mixing mostly focus on linguistics \cite{bokamba1988code}. This branch of work inspires a lot of computational efforts to model code-mixing based on syntactic constraints \citet{pratapa2018language}
However, code-mixing is also a social phenomenon, where the choice of language and the switching encode implicit meanings that only people within the same community would understand, and it often reflects intimacy and group identity \cite{ho2007code, kim2014sociolinguistic}. In order to model code-mixing sentences, one would need knowledge of the degree of multilinguality in the community, the speaker and the audience, their relationship, the occasion, and the intended effect of the communication \cite{bokamba1989there}.

%Different from \textit{code-switching}, the formal definition of \textit{code-mixing} is that in the process of speaking two languages (codes), a third one emerges, meaning that different aspects of the two languages are incorporated into a structurally assignable pattern \cite{al2018code}. Therefore, it is valid to view code-mixing languages as a different language from either of the parent languages, and that it's extremely low resource. In a code-mixing sentence consisting of two parent languages, one language is typically more ``dominant," controlling the syntax of the sentence, and the other language replaces some of the words or phrases. The dominant language is called a matrix language and the language that supply the phrase meaning is called the embedded language \cite{auer2005embedded, myers1992comparing}.
Code-mixing language is a different language from either of the parent languages and that it's extremely low-resource. In a code-mixing sentence consisting of two parent languages, one language is typically more ``dominant," controlling the syntax of the sentence, and the other language replaces some of the words or phrases. The dominant language is called a matrix language ($\mathcal{M}$) and the language that supplies the phrase meaning is called the embedded language ($\mathcal{E}$) \cite{auer2005embedded, myers1992comparing}.
Code-mixing can also happen at different levels. Intersentential code-mixing means the speaker mixes monolingual sentences together; intrasentential code-mixing is when words in the same sentence are from different languages, sometimes to express emphasis; and intra-word code-mixing, where inflection patterns or subword units are taken from different languages to form a word \cite{myers1989codeswitching}. In our work, we will focus on intrasentential code-mixing.

%mention that all data contain english

\subsection{Code Mixing NLP}
Code-mixing has gained growing interest in the \ac{nlp} community in recent years. Since it happens spontaneously during casual conversations, code-mixing is relatively more well-studied in the speech domain, such as automatic speech recognition \cite{chan2009automatic}. Previous work on text-based code-mixing \ac{nlp} has mostly focused on social media data \cite{thara2018code, bali2014borrowing, rudra2016understanding}. The spelling errors and script inconsistency in code-mixing languages make code-mixing \ac{nlp} a harder challenge on top of its low-resource nature. 
%English-Spanish and English-Hindi are among the most extensively studied language pairs because of the language diversity in these regions. 
Existing works have been attempting to improve the quality of code-mixing entity extraction \cite{rao2016cmee}, question answering \cite{obrocka2019prevalence}, and fine-tuning multilingual \ac{plm}s for intent prediction and slot filling \cite{krishnan2021multilingual}. Many NLP tools for code-mixing languages fine-tune multilingual models on scarce, manually labeled code-mixing data \citep{gautam2021translate}. Currently, available parallel corpora rely on non-scalable manual annotations, adding bias and noise to the data \citep{dhar-etal-2018-enabling, srivastava2020phinc}. Other applications involve using code-mixing data to finetune mBERT \cite{devlin2018mbert} to align multilingual embedding space \cite{qin2020cosda}.

\paragraph{Data Augmentation} \label{related:par:scm}
Some existing attempts to generate synthetic code-mixing data use subjective rule-based systems on parallel corpora, but standardized metrics like BLEU \citep{papineni2002bleu,doddington2002automatic} have proven them ineffective \citep{srivastava2021hinge}. \citet{pratapa2018language} used the Equivalence Constraint theory to force align the parse tree of the two languages to replace words in the source sentence. This approach results in very natural-sounding English-Hindi code-mixing sentences but relies on the assumption that parallel sentences in English and Hindi - both Indo-European languages - can be parsed with similar parse trees \cite{chang2015ancestry}, which is not often the case for more distant language pairs.

\paragraph{Sentiment Analysis}
Sentiment analysis is a crucial task in code-mixing \ac{nlp}. Such a model would be able to capture richer and more fine-grained sentiments in the switching of the language rather than the mere semantics of individual words. Some previous works code-mixing \ac{sa} follow a `translate-then-classify' paradigm, using labeled Hinglish and English parallel text to fine-tune mBART \cite{liu2020mbart} to translate the code-mixing sentences in Hinglish into English first, and then use a monolingual classifier to perform the SA task \citep{gautam2021translate}. Others have shown that fine-tuning BERT with natural code-mixing data achieves better results for downstream NLP tasks, albeit synthetic data helps with model responsivity \citep{santy2021bertologicomix}.

\subsection{Multilingual Pre-trained Models}\label{sec:plm}
Since code-mixing is a mixture of two or multiple languages, the use of multilingual language models is an important addition to code-mixing \ac{nlp}. In recent years, there have been a lot of transformer-based large pre-trained models trained on monolingual data from multiple languages in an attempt to capture multilingual information \cite{devlin2018mbert, conneau2019xlmr, liu2020mbart, xue2020mt5, ouyang2020erniem}. There has also been related work to train a multilingual language model specifically focus on the domain of social media. XLM-T is pre-trained on millions of tweets from over thirty languages, significantly outperforming its competitors on sentiment analysis and the TweetEval benchmark \cite{barbieri2022xlm}.

These models are trained on shared multilingual subword embeddings, but the context for training is still monolingual due to the nature of the training corpus. For example, the sentence embedding (not individual token) space of different languages in mBERT shows nearly no overlap, which limits its ability to understand code-mixing languages \cite{qin2020cosda}. Therefore, directly adopting multilingual language models trained on monolingual corpora from different languages is not enough for code-mixing \ac{nlp} due to its rich linguistic structure \cite{krishnan2021multilingual}. Often, natural code-mixing languages are not grammatically correct in either grammar of the two languages and lack syntactic constraints \cite{bokamba1989there}, making it harder to come up with a particular grammar for code-mixing based on the parent languages. 
    
\section{Methods}

\begin{figure*}[h!]
  \centering
  \includegraphics[width=0.95\textwidth,height=4cm]{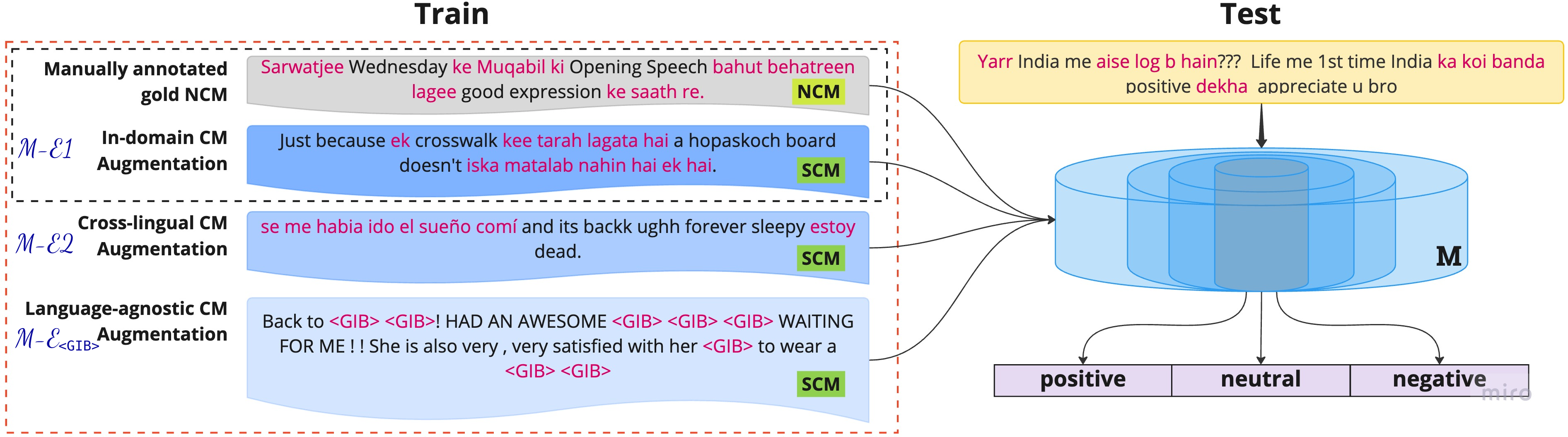}
  \caption{Different Synthetic Data Augmentation Methods. Inside the dotted black rectangular box are current data augmentation methods from the literature; inside the dotted red rectangular box is our novel data augmentation. The data (natural and synthetic) is used to fine-tune the \ac{sa} model $\mathcal{M}$, which outputs one of the three sentiment labels. }\label{fig:data}\vspace{-4mm}
\end{figure*}

In this section, we introduce a language-agnostic zero-cost data augmentation method that encodes the code-switch pattern with English and a constant mask, which provides an efficient universal data augmentation for any code-mixing sentences containing English.  Then, we propose three algorithms used to generate the \ac{scm}. 

\subsection{Language-Specific to Language-Agnostic}
\paragraph{Language-Specific}\label{sec:synthetic_cm}
First, we generate code-mixing sentences that are in the same language pair as the testing data.
We fix the matrix language ($\mathcal{M}$) in the generation process leveraging the language that has the most labeled monolingual data, and the other language in the pair is taken as the embedded language ($\mathcal{E}$), whose words and phrases we use to replace the phrases in $\mathcal{M}$. Labeled data in $\mathcal{M}$ can be fully utilized as we assume that code-mixing sentence generation from a monolingual sentence is label-preserving. We translate part of the source sentence in $\mathcal{M}$ into $\mathcal{E}$ and put the translation back to the source sentence to get the final \ac{scm} data, which is described more in detail in Section \ref{sec:generate}.
As shown in Figure \ref{fig:data} and described in Section \ref{related:par:scm}, this language-specific approach to code-mixing data augmentation has been adapted by previous methods. %cite?

\paragraph{Cross-lingual}
The domain of interest is social media data, which often contains code-mixing sentences that share the same matrix language, $\mathcal{M}$, and various embedded languages $\mathcal{E}i$ \cite{choudhury_2018}. Therefore, we investigate the effect of using code-mixing (both \ac{ncm} and \ac{scm}) in $\mathcal{M}$-$\mathcal{E}2$ as a cross-lingual data augmentation technique for \ac{sa} tasks on code-mixing sentences in $\mathcal{M}$-$\mathcal{E}1$. Similar to the Language-Specific method above, this approach leverages labeled data outside of the limited $\mathcal{M}$-$\mathcal{E}1$ domain. However, this uses cross-lingual \ac{ncm} data instead of monolingual data and can be used in conjunction with the language-specific \ac{scm} method described in Section \ref{sec:synthetic_cm}. %comeback

%Instead of using only labeled information in $\mathcal{M}$, the information contained in the labeled $\mathcal{M}$-$\mathcal{E}1$ data can be combined. 
%In our experiments, \ac{scm} sentences are generated with the matrix language being English and the embedded language being Hindi. We transliterate the Hindi from Devanagari script into the roman script to avoid any script discrepancy. We test this $\mathcal{M}$-$\mathcal{E}1$ \ac{scm} dataset on the labeled \textit{$\mathcal{M}$-$\mathcal{E}2$} \ac{ncm} \ac{sa} test set to evaluate the one-shot cross-lingual performance of our data augmentation.
\vspace{-1mm}
\paragraph{Language-Agnostic} 
Finally, we abandon any semantic information in $\mathcal{E}$ so the model focuses more on learning the pattern of code-switching rather than the semantics of individual words. To do this, we replace the embedded language tokens with a constant mask: \verb+<GIB>+, and create \ac{scm} datasets in the embedding-agnostic space $\mathcal{M}$-$\mathcal{E}_{<GIB>}$.
This data generation method is extremely low-cost, as the labeled \ac{sa} dataset in $\mathcal{M}$ is abundant, but more importantly, the translation into \verb+<GIB>+ is zero-cost - with no additional runtime or noise added during this trivial translation process.

\vspace{-1mm}
%\paragraph{{\color{red}{Paragraph Name}}}
\paragraph{}
As shown in Figure \ref{fig:data}, our method is the first to leverage cross-lingual $\mathcal{M}$-$\mathcal{E}2$ and language-agnostic $\mathcal{M}$-$\mathcal{E}_{<GIB>}$ data augmentation for code-mixing (dotted red box) in addition to the language-specific augmentation $\mathcal{M}$-$\mathcal{E}1$ (dotted black box). The synthetic data is mixed with \ac{ncm} in one-shot or without \ac{ncm} in zero-shot settings to train different \ac{sa} models in the second stage.
We evaluate the \ac{scm} datasets on the labeled Hinglish \ac{ncm} \ac{sa} test set of 3000 sentences, as well as different language pairs and across different low-resource natural training dataset sizes.

\subsection{Code-Mixing Generation (SCM)}\label{sec:generate}
%comeback
For both the language-specific and language-agnostic above, the data augmentation relies on effective \ac{scm} generation methods. 
%The majority of existing code-mixing generation methods use translation-related techniques such as end-to-end translation, lexical replacement, or bitext alignment.
In this section, we investigate two replacement-based algorithms to produce \ac{scm} sentences of any given language pair $\mathcal{M}$-$\mathcal{E}$.
Section \ref{sec:generate:random} introduces ways to create synthetic code-mixing sentences by replacing select tokens from the source sentence; in Section \ref{sec:generate:pos}, we take advantage of additional syntactic information to generate \ac{scm} by replacing phrases with \ac{pos} tagging.
%We investigate three different methods to create \ac{scm} data
%Our language-agnostic method also requires similar generation techniques.
%Due to the limited amount of \ac{ncm} data in the cross-lingual setting, we focus on using cross-lingual \ac{scm} data as a $\mathcal{M}$-$\mathcal{E}2$ data augmentation.
%three different \ac{scm} generation algorithms were used to produce synthetic data augmentation.

%There are multiple \ac{scm} generation algorithms that could be used for creating the code-mixing sentences in 
%The above-described methods all require designs to generate \ac{scm} sentences.

%We divide our code-mixing generation algorithm into three general sections, Section \ref{sec:generate:random} introduces ways to create synthetic code-mixing sentences by replacing select tokens from the source sentence; in Section \ref{sec:generate:pos}, we take advantage of additional syntactic information to generate \ac{scm} by replacing phrases with \ac{pos} tagging; Section \ref{sec:generate:ngram} trains an n-gram code-mixing language models on the \ac{ncm} training corpus and generate code-mixing sentences.

\subsubsection{Lexical Replacement}\label{sec:generate:random}
For each monolingual sentence in the matrix language $\mathcal{M}$, a word-level alignment translator translates select words into the embedded language $\mathcal{E}$ and a \ac{scm} sentence is generated by replacing those words from the matrix language to the embedded language. 

\paragraph{Code-Mixing Index}
For a given language pair $(\mathcal{M},\mathcal{E})$, we first calculate the \ac{cmi} \cite{gamback2014measuring} as follows:

\vspace{-4.5mm}
\begin{equation}
%     CMI=\begin{cases}
%  100\times(1-\frac{\max{(w_i)}}{n-u}),  & n>u \\
%   0, & n=u
% \end{cases}\vspace{-2mm}
CMI = 100(1-\frac{\max{(w_i)}}{n-u}) \text{ if } n>u \text{, else } 0\vspace{-2mm}
\end{equation}

where $w_i$ is the number of tokens for language $i$, $n$ is the total number of tokens and $u$ is the number of language-independent tokens. The \ac{cmi} measures the degree of code-mixing in a sentence when comparing different code-mixed corpora to each other. When selecting tokens from the source sentence, we match the number of tokens to the \ac{cmi} of the \ac{ncm} data so that the synthetic data has a similar distribution to the natural data. 

\paragraph{Token selection} As a strong \ac{scm} baseline, we randomly select tokens in the source sentence with probability equal to the \ac{cmi} of the \ac{ncm} corpus, and we name this replacement method \textit{random word replacement} inspired by \citet{krishnan2021multilingual}. Next, since switching points happens at the phrasal-level \cite{bokamba1989there}, we select random phrases in the source sentence to be replaced by the embedded language. This \textit{random phrase replacement} algorithm %is described in Algorithm \ref{alg:phrase}, in which $\tau$ is the temperature hyperparameter that we finetune to achieve desired \ac{cmi}.
is described more in deatil in Appendix \ref{app:alg1}.

% \begin{algorithm}
% \caption{Lexical Replacement}\label{alg:phrase}
% \KwData{$\mathcal S$}
% \KwResult{$\mathcal S_{cm}$}
% \SetKwRepeat{Do}{do}{while}
% \For{s \text{in} $\mathcal{S}$}{
%     $s' \gets \text{`'}$\;
%     $cr \gets 0$\;
%     \While{cr < \text{len}(s)}{
%         \eIf{random() < $\tau$}{
%             $L \gets rand(1,2,3)$\;
%             $phrase \gets \text{`'}$\;
%             $r \gets 0$\;
%             \For{$cr<\text{len}(s)$ and $r<L$}{ 
%                 phrase += s[curr]\;
%                 curr += 1\;
%                 r += 1\;
%             }
%             s' += translate(phrase)\;
%         }{
%         s' += s[cr]\;
%         cr += 1\;
%         }
%     }
%     $\mathcal S_{cm}$ += s'\;
% }
% \end{algorithm}

\begin{figure}[h!]
\centering\vspace{-2mm}
\includegraphics[width=0.49\textwidth]{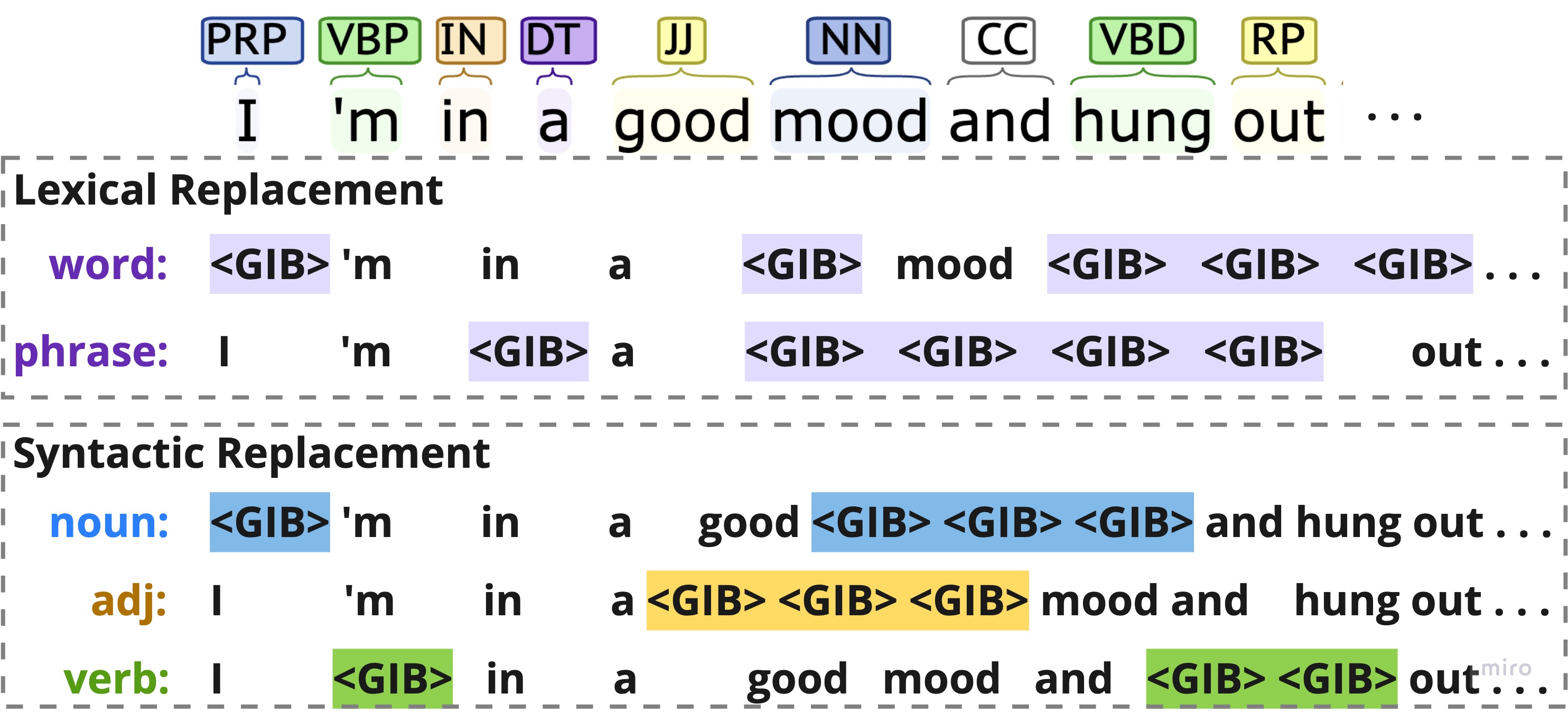}
\caption{\ac{scm} Generation\\
The lexical and syntactic algorithms to select parts of the sentence in $\mathcal{M}$ and replace them with tokens in $\mathcal{E}$.}\label{fig:scm_generate}\vspace{-3mm}
\end{figure}

\paragraph{Word level translation}
After selecting tokens in the source sentence, we use a word-level translation method to replace the tokens in $\mathcal{M}$ with tokens in $\mathcal{E}$ to generate the final \ac{scm} sentence. We investigate a word-level alignment method and fine-tuning mBART \cite{liu2020mbart} to translate the phrases as described more in detail in Appendix \ref{app:translation}. The translated phrase tokens are put back to the original position in the sentence of the matrix language to produce the final code-mixing sentence. Figure \ref{fig:scm_generate} shows the \ac{scm} generated with word and phrase-level lexical replacement of the language pair $\mathcal{M}$-$\mathcal{E}_{<GIB>}$.

% \begin{figure*}[h!]
%   \centering
%   \includegraphics[width=0.75\textwidth]{figures/generate.jpeg}
%   \caption{\ac{scm} Generation\\
% The lexical and syntactic algorithms to select parts of the sentence in $\mathcal{M}$ and replace them with tokens in $\mathcal{E}$.}\label{fig:scm_generate}\vspace{-3mm}
% \end{figure*}

\subsubsection{Syntactic Replacement}\label{sec:generate:pos}
Code-mixing is such a complex linguistic and social phenomenon that it is hard to come up with any universal syntactic constraints to formalize the switching of languages \cite{bokamba1989there}. Instead of randomly selecting words and phrases as in Section \ref{sec:generate:random}, we utilize more syntactic information to synthesize code-mixing sentences by tagging the \ac{pos} of each word in the source sentence. We use the Flair monolingual English \ac{pos} tagger \cite{akbik-etal-2018-contextual} with an accuracy of 97.85\%. Let $p$ be the pre-terminals from the Penn Treebank, our Syntactic Replacement Algorithm (more details in Appendix \ref{app:alg2}) returns a corpus $\mathcal{S}_p$, which is the dataset created by replacing all words with \ac{pos} tag $p$ from the source sentences with translation in the embedded language. And finally we take \begin{equation}
    \mathcal{S}_{cm}=\mathcal{S}_{p_1}\cup \mathcal{S}_{p_2}\cup\ldots\cup\mathcal{S}_{p_k}
\end{equation} as the combined dataset of all the above variations. We expect these to be more ``realistic" code-mixing sentences and thus more effective as a data augmentation. The lower half of Figure \ref{fig:scm_generate} shows the \ac{scm} generated with $p_i=$ noun, adj, and verb, respectively.

% \begin{algorithm}
% \caption{Syntactic Replacement}\label{alg:pos}
% \KwData{$\mathcal S$}
% \KwResult{$\mathcal S_{p}$}
% \SetKwRepeat{Do}{do}{while}
% \For{s \text{in} $\mathcal{S}$}{
%     $s' \gets \text{`'}$\;
%     \For{word \text{in} s}{
%         \eIf{\text{POS}(word) == p}{
%             word' $\gets$ translate(word)\;
%             s' += word'\;
%         }{
%         s' += word\;
%         }
%     }
%     $\mathcal S_{p}$ += s'\;
% }
% \end{algorithm}

\section{Experimental Setup}\label{sec:experiments}
\subsection{Datasets}\label{sec:exp:indomain}
Sentiment analysis is the main task of our data augmentation and we use both the accuracy and weighted f1 score (for three-way classification) to evaluate the results. The dataset used for training and testing is the SemEval-2020 Task 9 on Sentiment Analysis of Code-Mixed Tweets \cite{patwa2020semeval}\footnote{\url{https://ritual-uh.github.io/sentimix2020/}}, which consists of 14000, 3000, and 3000 Hinglish code-mixing sentences for training, we call this dataset Natural Code Mixing (NCM). 
Next, the source of the \ac{scm} generation is taken from the Stanford Sentiment140 dataset \cite{go2009twitter}
\footnote{\url{http://cs.stanford.edu/people/alecmgo/trainingandtestdata.zip}}, which consists of monolingual English tweets labeled with positive and negative sentiments. 
Finally, we use the English-Hindi bitext  from IITB \cite{kunchukuttan-etal-2018-iit} for the word alignment dictionary and for neutral \ac{scm} source sentences. 

To explore the cross-lingual generalizability of our data augmentation, we use the Spanglish data in the SemEval dataset (train: 9002, eval: 3000, test: 3000) and an English-Malayalam code-mixing dataset (train: 3000, eval: 1452, test: 1000) labeled for \ac{sa} \cite{chakravarthi2020malayalam} as cross-lingual evaluation test sets. 
%Because the published test set for Spanglish \ac{ncm} does not contain labels for us to report accuracy, we split the published training set into 9002 training sentences and 3000 test sentences. 
%We also use an English-Malayalam code-mixing dataset labeled for sentiment analysis \cite{chakravarthi2020malayalam}. The dataset contains 5452 labeled sentences and we divide it into 3000, 1452, and 1000 sentences for training, validation, and testing, respectively. 

\subsection{Preprocessing}\label{sec:exp:preprocess}
A coarse filtering \cite{liu2020kk2018} of the \ac{ncm} data is performed before they are fed into the language model, as we want to take out the tokens unknown from the language model but preserve most of the original style of the tweets. First, empty strings, hash symbols, and URLs are removed from the text. All emojis and emoticons are replaced by their English descriptions using the emoji library\footnote{\url{https://pypi.org/project/emoji/}}.

For the data augmentation, we perform the same filtering method on the Sentiment140 English dataset. Since the Sentiment140 dataset only contains binary sentiments, we use roBERTa-base finetuned on English Twitter \ac{sa} \cite{barbieri2020tweeteval} to mine neutral sentences in English from the IITB parallel English-Hindi corpus. During mining, we collect sentences that are classified as neutral with a confidence score higher than 0.85 into the final source language dataset.

\subsection{Training}
To make the data augmentation more effective, we adopt a gradual fine-tuning approach \cite{xu2021gradual}. Treating the \ac{scm} data as out-of-domain data, since it does not have the exact distribution as the human-produced \ac{ncm} sentences, we iteratively fine-tune on the mixed \ac{scm} and \ac{ncm} with decreasing amounts of out-of-domain data. This gradual fine-tuning approach allows the model to better fit the distribution of the target domain, as the training data gets more and more similar to the domain of the test data, which is \ac{ncm} in our case. We fine-tune the model in 5 stages with the amount of \ac{scm} data size of $[30000,10000,3000,1000,0]$ with 3 epochs in each stage. An embedding length of 56 is used for the Hinglish corpus and 40 for the Spanglish corpus following the baseline method in \citet{patwa2020semeval}. The AdamW optimizer is used with a linear scheduler and a learning rate of 4e-6, which is determined empirically by preliminary experimentation.

\subsection{Multilingual \ac{plm}s}
In order to stay consistent with the baseline accuracy provided in \citet{patwa2020semeval}, we primarily evaluate our synthetic data augmentation on mBERT with a transformer classifier, but we also explore different models (XLM-R and XLM-T) in our ablation studies. As shown in Figure \ref{fig:data}, our method takes inspiration from previous work to fine-tune a multilingual \ac{plm} with a combination of natural and synthetic data of the same language pair, then we expand into cross-lingual and language-agnostic data augmentation. In order to directly compare the effectiveness of different \ac{scm} datasets, we keep the size of the augmented data consistent throughout our experiments at 30000 sentences and the size of the natural data consistent at 3000 sentences. We evaluate our data augmentation on the labeled Hinglish \ac{ncm} \ac{sa} test set of 3000 sentences and repeat all experiments for 5 trials to report the standard deviation.

\subsection{Baselines and Hyperparameters}
In order to highlight the effect of our data augmentation method, we use the same model used in the shared task baseline \cite{patwa2020semeval} and we were able to reproduce the baseline f1 score of $0.65$. The model uses mBERT and a transformer classification layer to predict the sentiment to be positive, negative, or neutral. To demonstrate the effectiveness of our data augmentation in low-resource settings, we cut down the training dataset to 3000 sentences for all future experiments.

In our 5-stage gradual fine-tuning procedure, the \ac{ncm} data is passed through the model $5\times3=15$ times (3 epochs per stage), so we consider two candidates for the baseline model without data augmentation. We fine-tune the model for 3 epochs (1 stage) and 15 epochs (5 stages) with only \ac{ncm} and compare the model performance. As shown in exp \#1 and \#12 in Table \ref{tab:enhi_scm}, the two procedures produced results that are not statistically different, and a converging loss curve indicates that the 5-stage fine-tuning will lead to model over-fitting to the training dataset. Therefore, we use the one-stage fine-tuning on the \ac{ncm} as a strong baseline for comparisons.

\begin{figure}[h!]
\centering\vspace{-3mm}
\includegraphics[width=0.44\textwidth]{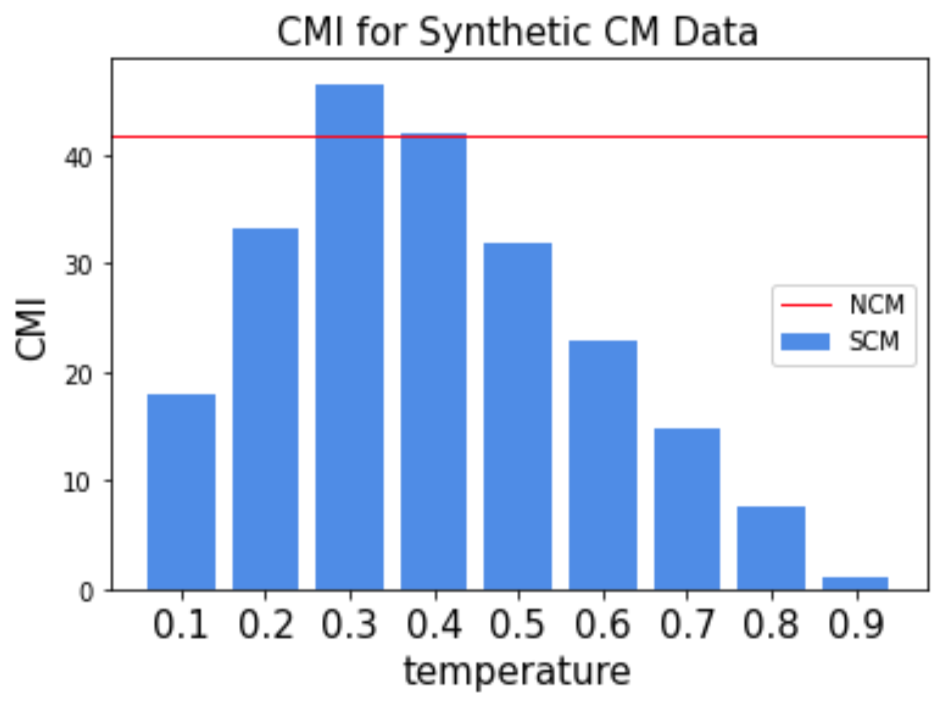}\vspace{-1mm}
\caption{Temperature Tuning for \ac{cmi}\\
When $\tau=0.4$, the synthetic data most closely resembles natural code mixing data (red line) in terms of \ac{cmi} }\label{fig:temperature}
\end{figure}

Figure \ref{fig:temperature} shows the \ac{cmi} of the synthetic data generated by the phrase level replacement algorithm using different temperature values $\tau$. Upon further observation, the natural data contains more Hindi tokens than English tokens, so a $\tau$ value to the right of the peak should be chosen if we use English as the source language in the generation algorithm because a higher temperature means more tokens will be selected to be translated into the embedded language. When $\tau=0.4$, the \ac{cmi} of the \ac{scm} is the closest to the \ac{cmi} of the \ac{ncm}.

\section{Results and Analysis}

We first compare the language-specific, cross-lingual, and language-agnostic augmentation results in Section \ref{sec:results:all}, and then discuss multiple strong baselines and compare different lexical replacement variants in Section \ref{sec:results:lexical}. In the ablation studies, we examine different syntactic replacement variants, effectiveness across low-resource settings, and the generalizability of our \ac{scm} across different \ac{plm}s.

\subsection{\ac{scm} Data Augmentation}\label{sec:results:all}
There are two \ac{scm} datasets being evaluated - English-Hindi \ac{scm} generated using the phrase level lexical replacement algorithm (hiSCM) and English-XX \ac{scm} generated using the syntactic replacement algorithm (gibSCM). In addition to the English-Hindi \ac{ncm}, we test on two previously unseen English-Spanish and English-Malayalam \ac{ncm} datasets. Spanish and English have a small Levenshtein distance \cite{serva2008indo}, while Malayalam and English come from different language families. Additionally, it is important to note that although both Hindi and Malayalam are widely spoken in India, Hindi is in the Indo-European language family whereas Malayalam is from the Dravidian language family \cite{emeneau1967south}, so Malayam is a lot more different from English than Hindi.

First, the baselines in rows 1, 4, 7 of Table \ref{tab:main} are the weighted f1 scores obtained by fine-tuning mBERT on \textit{only} the \ac{ncm} dataset. Fine-tuning the combined English-Hindi \ac{ncm} and \ac{scm} show our in-domain language-specific data augmentation as shown in exp \#2. Then, we have the cross-lingual experiments that uses English-Hindi \ac{scm} as a data augmentation for English-Spanish and English-Malayalam tasks as shown in exp \#5 and \#8, respectively. Finally, our novel language-agnostic augmentation of masked English-XX code-mixing is tested with all language pairs as shown in exp \#3, \#6, and \#9.

As shown in Table \ref{tab:main}, the synthetic English-Hindi corpus is extremely helpful for all language pairs. It is important to note that the language-agnostic gib\ac{scm} augmentation (exp \#6 and \#9) achieved better performance compared to the English-Hindi \ac{scm} augmentation (exp \#5 and \#8) on the cross-lingual datasets of English-Spanish and English-Malayalam. This implies that Hindi phrases in the English-Hindi \ac{scm} might be biasing the model during training, making the language-agnostic data a more powerful universal data augmentation.

\begin{table}[h!]
\centering
\begin{tabular}{clcccc}
\hline \textbf{\#} & \textbf{Language} & \textbf{Data} & \textbf{SCM} & \textbf{$\uparrow$ (\%)}\\ \hline
1 & Hindi & hi\ac{ncm}     & 0.5505 & 0  \\
2 & Hindi & +hi\ac{scm}    & \textbf{0.5860} & 6.45  \\
3 & Hindi & +gib\ac{scm}   & 0.5853 & 6.32 \\
\hline
4 & Spanish & es\ac{ncm}   & 0.4956 & 0  \\
5 & Spanish & +hi\ac{scm}  & 0.5053 & 1.96  \\
6 & Spanish & +gib\ac{scm} & \textbf{0.5061} & 2.12  \\
\hline
7 & Malayalam & ml\ac{ncm}     & 0.6703 & 0  \\
8 & Malayalam & +hi\ac{scm}   & 0.7202 & 7.45 \\
9 & Malayalam & +gib\ac{scm}  & \textbf{0.7221} & 7.73  \\
\hline
\end{tabular}
\cprotect\caption{\label{tab:main} Cross-Lingual Data Augmentation\\
Synthetic data generated by translating select tokens into either Hindi or the \verb|<GIB>| mask. `+' means combining synthetic data with \ac{ncm} for gradual fine-tuning.}
\end{table}

\subsection{English-Hindi Lexical Replacement}\label{sec:results:lexical}
We evaluate the data augmentation generated using the Lexical Replacement Algorithm (Algorithm \ref{alg:phrase}) with the matrix language being English and the embedded language being Hindi. %and the results are shown in Table \ref{tab:enhi_scm}. 
When there are zero available labeled natural data, fine-tuning on \textit{only} the synthetic data (exp \#11) achieved a higher weighted f1 score than the zero-shot performance (exp \#10) on the un-fine-tuned mBERT. This indicates that the \ac{scm} is an effective zero resource data augmentation for code-mixing \ac{sa}. 
Furthermore, the more complex phrase-level lexical replacement algorithm (exp \#2) achieves a higher improvement compared to the naive random word lexical replacement (exp \#13). This supports our intuition that code-mixing is more complex than just a random combination of words from different languages, in which a bilingual speaker just picks any word that comes to mind. Rather, intrasentential code-mixing happens at the phrase level.

\begin{table}[h!]
\centering
\begin{tabular}{ccc}
\hline \textbf{\#} & \textbf{Experiment} & \textbf{Weighted F1}\\ \hline
10 & zero-shot MBERT               & 0.1696 $\pm$ 0.0248  \\
11 & zero-shot \ac{scm}      & 0.3144 $\pm$ 0.0787  \\
1 & baseline (one-stage)    & 0.5505 $\pm$ 0.0041  \\
12 & baseline (five-stage)   & 0.5518 $\pm$ 0.0075  \\
13 & +hi\ac{scm} (word)   & 0.5719 $\pm$ 0.0099  \\
2 & +hi\ac{scm} (phrase) & \textbf{0.5860} $\pm$ 0.0112  \\
\hline
\end{tabular}
\caption{\label{tab:enhi_scm} Hinglish Lexical Replacement\\
Synthetic data made by translating select words/phrases with translation from English to Hindi. `+' means combining synthetic data with \ac{ncm} for gradual fine-tuning.}\vspace{-3mm}
\end{table}

\subsection{Ablation Studies}
\subsubsection{English-XX Syntactic Replacement}\label{sec:results:pos}
After observing the cross-lingual effectiveness of our \ac{scm} algorithm, we attempt to remove the semantic information of the embedded language completely by using a constant \verb+<GIB>+ mask in place of translated Hindi tokens. This way, the model learns the pattern of intrasentential code-switching rather than simply gathering the semantics of the lexicons. Additionally, this reduce the cost of \ac{scm} generation as no translation process is involved. Table \ref{tab:gibpos} shows the classification accuracy with English-XX \ac{scm} datasets generated with the Lexical Replacement Algorithm, Ngram Generator, and the Syntactic Replacement Algorithm with different sub-strategies and their performance. The syntactic replacement algorithm with all nouns ``translated" into \verb+<GIB>+ outperformed the baseline and all other \ac{scm} algorithms, achieving a 6.32\% relative improvement over the strong baseline performance. This provides a language agnostic data augmentation method for any code-mixing languages with English as the matrix language, which is the most common language in code-mixing sentence pairs in social media \cite{thara2018code}.

\begin{table}[h!]
\centering
\begin{tabular}{clccc}
\hline \textbf{\#} & \textbf{Exp.} & \textbf{f1} & \textbf{$\uparrow$ (\%)}\\ \hline
1 & Baseline (one-stage)            & 0.5505 & 0 \\
14 & Lexical (word)      & 0.5686 & 3.29 \\
15 & Lexical (phrase)    & 0.5664 & 2.88 \\
16 & Ngram (combined)    & 0.5540 & 0.63 \\
17 & Syntactic (Adj)     & 0.5633 & 2.33  \\
18 & Syntactic (Verb)    & 0.5723 & 3.97  \\
19 & Syntactic (Noun)    & 0.5786 & 5.10  \\
3 & Syntactic (mixed)   & \textbf{0.5853} & \textbf{6.32}  \\
\hline
\end{tabular}
\cprotect\caption{\label{tab:gibpos} \ac{scm} Generation Algorithms\\ Synthetic English-\verb|<GIB>| \ac{scm} generated by lexical and syntactic strategies with slightly different token selection methods.}\vspace{-3mm}
\end{table}

\subsubsection{Low Resource Levels}
\begin{figure}[h!]
\includegraphics[width=0.47\textwidth]{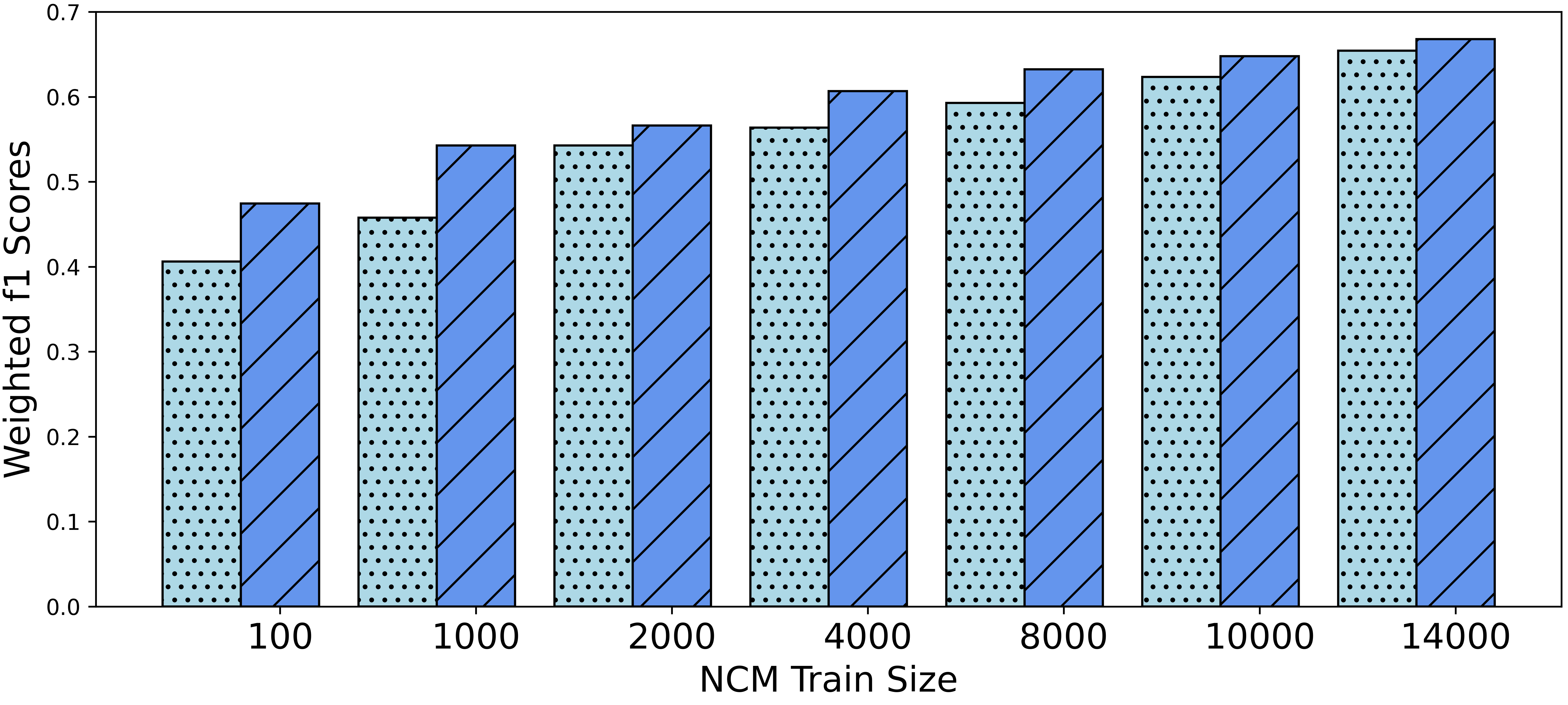}
\caption{Effect of \ac{scm} Across Low Resource Levels\\
Weighted f1 scores with \ac{scm} English-XX Lexical Replacement data augmentation across different low-resource levels. The \ac{scm} is more effective with decreased natural dataset size.}\label{fig:lowre}\vspace{-4mm}
\end{figure}

The difficulty to collect code-mixing data, especially in the text form, makes this prevalent linguistic phenomenon common in real life yet scarce in \ac{nlp} research. We artificially limit the size of training data to create extremely low-resource scenarios, while keeping the test set constant to evaluate the effect of \ac{scm} augmentation in low-resource settings. Figure \ref{fig:lowre} shows the weighted f1 scores for \ac{sa} on the \ac{ncm} test set of 3000 sentences when trained on varying amounts of \ac{ncm} training data, while the amount of \ac{scm} augmented data in each stage of the gradual fine-tuning is kept constant.
The \ac{scm} data augmentation is helpful for training across all low-resource levels. As the size of natural training data becomes more and more limited, our data augmentation becomes significantly more effective. The f1 scores of this experiment can be found in Appendix \ref{app:lowre}.

In the extremely low resource case (100 \ac{ncm} sentences), fine-tuning the model with \ac{scm} improves the f1 score from 0.3062 to 0.4743, resulting in a 54.9\% relative improvement. As the amount of available \ac{ncm} training data becomes available, both the baseline and the data-augmented models become better at predicting the correct label. The data augmentation on the largest training set of 14000 sentences produced the best results. 

Fine-tuning with the large \ac{ncm} corpus of 14000 sentences achieves a weighted f1 score of 0.6543, which is higher than the \ac{scm} data augmentation in the extreme low-resource scenario. This shows that human-produced gold data is still undoubtedly superior to the synthetic data, but synthetic data is still helpful in the absence of natural data.
We see a smoother loss curve during training when the \ac{ncm} data size is large (i.e. when the \ac{scm}/\ac{ncm} ratio is smaller). This is due to the domain consistency of the \ac{ncm} data. This indicates further opportunities to stabilize the synthetic data style so that the model's search path has less noise.

\subsubsection{Generalizability Across Models}
We use our data augmentation method on more powerful models to evaluate the generalizablity of the \ac{scm} corpus across different models. Table \ref{tab:models} shows the baseline of one-shot fine-tuning on \ac{ncm} vs. \ac{scm} performance of mBERT, XLM-R, and XLM-T on the Hinglish \ac{sa} task. XLM-R performs extremely well on low-resource languages \cite{conneau2019xlmr}, and XLM-T is an XLM-R-based model pre-trained on millions of tweets in over thirty languages \cite{barbieri2022xlm}.

\begin{table}[h!]
\centering
\scalebox{0.9}{
\begin{tabular}{clcccc}
\hline \textbf{\#} & \textbf{Model} & \textbf{Baseline} & \textbf{+gib\ac{scm}} & \textbf{$\uparrow$ (\%)}\\ \hline
1  & mBERT   & 0.5505 & 0.5853  & 6.32  \\
%19 & RoBERTa & 0.2846 &  0.5965 & 109.6 \\
20 & RoBERTa-T & 0.5982 & 0.6274  & 4.88 \\
21 & XLM-R   & 0.6034 & 0.6564  & 5.30 \\
22 & XLM-T   & 0.6491 & 0.6600  & 1.68  \\
\hline
\end{tabular}
}
\caption{\label{tab:models} Generalizability across Models\\
gib\ac{scm} focuses on teaching code-switch patterns; it outperform one-shot \ac{ncm} baselines on all \ac{plm}s.}\vspace{-5mm}
\end{table}

The data augmentation improves model performance for all pre-trained multilingual language models. This indicates that even powerful multilingual \ac{plm}s are not able to completely capture the complex structure of code-mixing languages. Although the improvement becomes smaller as the model gets more complex, the data augmentation methods consistently outperform the strong baseline. The smaller improvement of our data augmentation technique on XLM-R and XLM-T might also result from the fact that they are trained on Common-Crawl and Twitter data, which contains code-mixing sentences occasionally. And XLM-T has already seen a large amount of Twitter data. Overall, the synthetic data mimics the pattern of code-switching and helps the model adjust to the in-domain code-mixing training data.

\subsubsection{Other Generation Methods}\label{sec:generate:ngram}\vspace{-0.5mm}
%\subsubsection{N-gram Code-Mixing Language Model}\label{sec:generate:ngram}
Note that all our methods above rely on replacement-based generation methods. However, there are other data augmentation techniques.
Our work focuses on the cross-lingual and language-agnostic nature of the \ac{scm} rather than investigating the optimal data augmentation algorithms. In preliminary experiments, we were able to demonstrate the language-agnostic design also works on other statistical and neural data generation techniques such as n-gram language modeling with positive improvement on the \ac{sa} task, demonstrating that the language choice, including $\mathcal{M}$-$\mathcal{E}_{<GIB>}$, is technique-agnostic. We describe some of these techniques in Appendix \ref{app:ngram}.

%ngram, tree alignment, gpt (autoregressive monolingual pre-trained language models), etc.
%move to appendix
%We train n-gram language models on the limited \ac{ncm} data with the assumption that a model trained on positively labeled sentences will be a ``positive model" that generates positive code-mixing sentences. The generation capability is limited by the small training corpus size, and there is a trade-off between $n$ in an n-gram language model and the diversity of the sentence generated. Therefore, for each of the positive, neutral, and negative subsets of the training corpus, we train trigram, 4-gram, 5-gram, and 6-gram language models with backoff and add-lambda smoothing. We combine the generated sentences to take advantage of both the quality and diversity of the generated data.

\section{Conclusion}\label{sec:conclusion}\vspace{-1.5mm}
In this work, we introduce two replacement-based algorithms for synthetic code-mixing for training data augmentation. Most importantly, we prove a low-cost, language-agnostic solution to the data scarcity problem of code-mixing corpus for \ac{nlp} tasks, specifically \ac{sa}. Analyzing the sentiment analysis performance of the \ac{scm} as a data augmentation gives insight into the phenomenon of code-mixing.
%\paragraph{Future Directions}
Our algorithm is flexible enough to be easily extended to any other replacement strategies, making it a universal framework for future explorations of the pattern of code-mixing languages. %Moreover, our currently proposed algorithm is language agnostic to the embedded language. This can easily be extended to matrix language agnostic by either naively flipping the token selection or training a keyword extraction model to and replacing the non-keywords with a gibberish mask, in which case more careful linguistic design would be necessary.

\section*{Acknowledgments}\label{sec:acknowledgement}This work was supported, in part, by the Human Language Technology Center of Excellence (HLTCOE) at Johns Hopkins University.

%\clearpage
\bibliography{custom}

\clearpage
\appendix

\section{Lexical Replacement Algorithm}\label{app:alg1}
\begin{algorithm}[h!]
\caption{Lexical Replacement}\label{alg:phrase}
\KwData{$\mathcal S$}
\KwResult{$\mathcal S_{cm}$}
\SetKwRepeat{Do}{do}{while}
\For{s \text{in} $\mathcal{S}$}{
    $s' \gets \text{`'}$\;
    $cr \gets 0$\;
    \While{cr < \text{len}(s)}{
        \eIf{random() < $\tau$}{
            $L \gets rand(1,2,3)$\;
            $phrase \gets \text{`'}$\;
            $r \gets 0$\;
            \For{$cr<\text{len}(s)$ and $r<L$}{ 
                phrase += s[curr]\;
                curr += 1\;
                r += 1\;
            }
            s' += translate(phrase)\;
        }{
        s' += s[cr]\;
        cr += 1\;
        }
    }
    $\mathcal S_{cm}$ += s'\;
}
\end{algorithm}

\section{Syntactic Replacement Algorithm}\label{app:alg2}
\begin{algorithm}[h!]
\caption{Syntactic Replacement}\label{alg:pos}
\KwData{$\mathcal S$}
\KwResult{$\mathcal S_{p}$}
\SetKwRepeat{Do}{do}{while}
\For{s \text{in} $\mathcal{S}$}{
    $s' \gets \text{`'}$\;
    \For{word \text{in} s}{
        \eIf{\text{POS}(word) == p}{
            word' $\gets$ translate(word)\;
            s' += word'\;
        }{
        s' += word\;
        }
    }
    $\mathcal S_{p}$ += s'\;
}
\end{algorithm}

\section{Phrase-Level Translation}\label{app:translation}
We experiment with two different word-level translation methods in this section. First, we create a simple one-to-many weighted English-Hindi dictionary using Awesome-Align \cite{dou2021word}\footnote{\url{https://pypi.org/project/awesome-align/}}. Take the Hinglish code-mixing, for example, we generate word-level alignments using an English-Hindi parallel corpus. For each English token, the aligned Hindi word is collected to create a weighted list. For the tokens to be replaced in the source sentence, an aligned Hindi word is randomly selected from the weighted word-level dictionary. 
The second translation method uses mBART \cite{liu2020mbart} to translate the English phrases into Hindi. 

\section{Low Resource Levels}\label{app:lowre}

\begin{table}[h!]
\centering
\begin{tabular}{clcccc}
\hline \textbf{\#} & \textbf{|\ac{ncm}|} & \textbf{Baseline} & \textbf{+gib\ac{scm}} & \textbf{$\uparrow$ (\%)}\\ \hline
21 & 14000 & 0.6543 & \textbf{0.6679} & 2.08 \\
22 & 12000 & 0.6370 & 0.6613 & 3.81 \\
23 & 10000 & 0.6234 & 0.6477 & 3.89 \\
24 & 8000  & 0.5929 & 0.6322 & 6.63 \\
25 & 6000  & 0.5782 & 0.6152 & 6.40 \\
26 & 4000  & 0.5638 & 0.6068 & 7.63 \\
5  & 3000  & 0.5505 & 0.5853 & 6.32 \\
27 & 2000  & 0.5426 & 0.5662 & 4.35 \\
28 & 1000  & 0.3578 & 0.5427 & 33.08 \\
29 & 500   & 0.3130 & 0.5259 & 48.98 \\
30 & 100   & 0.3062 & 0.4743 & \textbf{54.90} \\
\hline
\end{tabular}
\caption{\label{tab:lowre} Effect of \ac{scm} Across Low Resource Levels}
\end{table}

\section{N-gram \ac{scm} Generation}\label{app:ngram}
We train n-gram language models on the limited \ac{ncm} data with the assumption that a model trained on positively labeled sentences will be a ``positive model" that generates positive code-mixing sentences. The generation capability is limited by the small training corpus size, and there is a trade-off between $n$ in an n-gram language model and the diversity of the sentence generated. Therefore, for each of the positive, neutral, and negative subsets of the training corpus, we train trigram, 4-gram, 5-gram, and 6-gram language models with backoff and add-lambda smoothing. We combine the generated sentences to take advantage of both the quality and diversity of the generated data.

\end{document}